\documentclass{article}

\usepackage[preprint]{corl_2023} 

\usepackage[ruled,vlined]{algorithm2e}

\usepackage[dvipsnames]{xcolor} 
\usepackage{graphicx}
\usepackage{textcomp}
\usepackage{hyperref}

\usepackage{tkz-graph}
\usetikzlibrary{shapes.geometric}

\usepackage{tikz}
\usetikzlibrary{decorations.pathreplacing}

\usepackage{array}
\usepackage{tabularx}
\usepackage{subcaption} 

\newcolumntype{b}{X}
\newcolumntype{s}{>{\hsize=.4\hsize}X}
\newcolumntype{m}{>{\hsize=.7\hsize}X}
\newcolumntype{t}{>{\hsize=.3\hsize}X}
\newcolumntype{v}{>{\hsize=.2\hsize}X}
\newcolumntype{q}{>{\hsize=.1\hsize}X}

\usepackage{makecell}

\usepackage{amsfonts}

\usepackage{amsmath}
\usepackage{amssymb}
\DeclareMathOperator*{\argmax}{arg\,max}

\usepackage{float}

\usepackage{threeparttable}
\usepackage{soul}

\usepackage{listings}
\usepackage{fancyvrb}
\usepackage{multirow} 

\usepackage{pifont}

\makeatletter
\newcommand\notsotiny{\@setfontsize\notsotiny\@vipt\@viipt}
\makeatother

\title{Integrating LLMs and Decision Transformers for Language Grounded Generative Quality-Diversity}

\author{
  Achkan Salehi\\
  ISIR/CNRS\\
  Sorbonne University\\
  \texttt{achkan.salehi@sorbonne-universite.fr} \\
  \And
  Stephane Doncieux\\
  ISIR/CNRS\\
  Sorbonne University\\
  \texttt{stephane.doncieux@sorbonne-universite.fr} \\
}

\begin{document}
\maketitle

\begin{small}

\begin{abstract}
  Quality-Diversity is a branch of stochastic optimization that is often applied to problems from the Reinforcement Learning and control domains in order to construct repertoires of well-performing policies/skills that exhibit diversity with respect to a behavior space. Such archives are usually composed of a finite number of reactive agents which are each associated to a unique behavior descriptor, and instantiating behavior descriptors outside of that coarsely discretized space is not straight-forward. While a few recent works suggest solutions to that issue, the trajectory that is generated is not easily customizable beyond the specification of a target behavior descriptor. We propose to jointly solve those problems in environments where semantic information about static scene elements is available by leveraging a Large Language Model to augment the repertoire with natural language descriptions of trajectories, and training a policy conditioned on those descriptions. Thus, our method allows a user to not only specify an arbitrary target behavior descriptor, but also provide the model with a high-level textual prompt to shape the generated trajectory. We also propose and LLM-based approach to evaluating the performance of such generative agents. Furthermore, we develop a benchmark based on simulated robot navigation in a 2d maze that we use for experimental validation. The code for this paper is available at \texttt{https://github.com/salehiac/LanguageGroundedQD}.
\end{abstract}

\section{Introduction}
  \label{sec_intro}
 
  Quality-Diversity (QD) optimization \cite{pugh2016quality} is a branch of stochastic optimization which primarily aims at finding diverse and locally well-performing solutions, where the notion of locality is tied to a behavior descriptor space. Although based on conceptually simple evolutionary principles, QD optimization has proved powerful in increasing robustness through discovering high-reward solutions in different behavioral niches \cite{cully2015robots} as well as in decision making problems that are challenging for traditional Reinforcement Learning (RL) methods, such as sparse rewards situations in which exploration is indispensable \cite{ecoffet2019go, lehman2011abandoning, kim2021exploration, cully2015robots, mouret2015illuminating, huber2022e2r}.

  An issue that limits the applicability of QD methods is the discrete and finite nature of the repertoire of solutions that they build. What if the exact behavior that we want to sample is not already included in the archive? What if we want to customize the behavior beyond what is possible with a single behavior descriptor? A possible solution to the first question is given by the early work of Jegorova \textit{et al.}\cite{jegorova2020behavioral}, in which a generative model conditioned on the target behavior descriptor is trained using the contents of a repertoire of open-loop policies. More recently, Mace \textit{et al.} have proposed the QD-transformer\cite{mace2023quality}, which conditions a transformer on the desired behavior descriptor. None of those works, however, address the second question.

  In this work, we suggest the use of natural language to further customize the behavior of agents that are sampled. Specifically, our aim is as follows: given a target behavior descriptor, we want to sample a policy that not only exhibits that behavior, but that is able to adapt it to a natural language command. For example, given a navigation task where the robot is given a target behavior descriptor in the form of $2d$ coordinates, we also want the policy to follow high-level guidelines such as "Can you go from the table to the door? You should circle the plant but avoid getting close to the carpet". We propose a solution based on leveraging LLMs to add textual descriptions to QD repertoires which will serve as a basis for training a causal transformer simultaneously conditioned on behavior descriptors and tokens from commands/descriptions given in natural language. As our aim is to develop a method that is applicable to repertoires constructed for robotics control tasks, we focus on continuous action spaces. In this paper, We also discuss different possibilities for evaluating the performance of those models, and propose an LLM-based approach which requires little in terms of labeling while remaining reasonably aligned with evaluations made by humans. 

  To evaluate the proposed approach, we have developed a benchmark based on maze navigation that follows the desiderata\footnote{According to those, a benchmark should be challenging, targeted and scalable, and should allow rapid experiments and iteration over solutions.} stated in previous works \cite{osband2019behaviour, salehi2022towards}. It is essentially an extension to the maze navigation benchmarks that are popular in the QD literature which results from adding several sprites representing household objects as well as colored tiles to a randomly generated maze. We note that our simulated environment guarantees reproducibility, and that it therefore does not require uncertainty mitigation strategies. This is a useful simplification that allows us to focus on model architecture.

  Our code is available at \texttt{https://github.com/salehiac/LanguageGroundedQD}.

  The paper is organized as follows. The next section (\S\ref{sec_formulation}) is dedicated to preliminaries, notations and formal problem statements. The proposed method is detailed in \S\ref{sec_proposed_method}, and metrics for evaluating such models are discussed and defined in \S\ref{sec_how_to_eval}. Experiments and their results are detailed in S\ref{sec_exp}. Positioning with respect to related works is the subject of \S\ref{sec_related_works}. This is followed by a discussion about limitations and perspectives \ref{sec_discussion}. Note that implementation details are reported in the appendix.

\section{Problem formulation}
  \label{sec_formulation}

  We consider decision processes in partially observable environments. Formally, noting $\mathcal{S}$ the state space and $\mathcal{O}$ the set of possible observations, we define $\Phi: S \times \mathcal{O} \rightarrow [0,1]$ as the probability $p(o|s)$ of an element $o \in \mathcal{O}$ given a state $s$. Let $\mathcal{A}$ and $r(.)$ respectively denote the action space and a reward signal. A Partially Observable Markov Decision Processes (POMDP) is then given by $<S,\mathcal{A},\mathcal{F},r(.),\mathcal{O}, \Phi,\gamma>$, where $\mathcal{F}:S \times S \times \mathcal{A} \rightarrow [0,1]$ is the transition probability function, \textit{i.e.} $\mathcal{F}(s',s,a)=P(s'|s,a)$, and where $\gamma \in [0,1]$ is a discount factor. Policies interacting with the environments that we consider will be written $\pi_{\theta}$ where $\theta$ is a set of learnable parameters residing in the parameter space $\Theta$. We consider the case in which observations and actions are made at discrete, regular timesteps, and use $\tau \triangleq \{o_0, a_0, ..., o_{H-1}, a_{H-1}, o_H\}$ to refer to observation-action space trajectories. When necessary for disambiguation, the notation $\tau(\pi)$ will be used to specify that the trajectory has been obtained with policy $\pi$. Note that we focus on continuous actions, which are more common in robotic control tasks than discrete commands.

  In contrast with the classical RL problem where the objective is to obtain a single agent that maximizes the expected (discounted) cumulative reward, we consider Quality-Diversity optimization where the aim is to build a repertoire $\mathcal{R}=\{\theta_i\}_i$ of diverse but locally well-performing agents. Those notions of diversity and locality are tied to a behavior descriptor space $\mathfrak{B}$, which along with a behavior descriptor function $\phi: \Theta \rightarrow \mathfrak{B}$ mapping each policy to an element of the behavior space, is assumed given. In essence, QD methods are evolutionary processes which view the parameter space $\Theta$ as a genotype space, and starting from a random population of agents $\mathcal{U}\subset \Theta$, define mutation\footnote{Cross-over operations between agents can also be used, but are much less frequently encountered in the QD-RL literature.}, evaluation and selection processes that (explicitly or implicitly) perform multi-objective optimization over the population, resulting in divergent behavior space exploration as well as reward maximization. For example, Novelty-Search\cite{lehman2011abandoning} based methods iterate over those steps at each generation: 1) mutation, which results in a population of offsprings $\mathcal{U}'$, 2) Fitness (\textit{i.e.} reward/return) and Novelty evaluation of $\mathcal{U} \cup \mathcal{U}'$, 3) Selection of the next population (\textit{e.g.} through Multi-objective optimization such as NSGAII), 4) Insertion of significantly well-performing or novel agents into the repertoire. QD algorithms base on MAP-elites \cite{mouret2015illuminating}, on the other hand, equate $\mathcal{R}$ and the population $\mathcal{U}$, and generally iterate through 1) mutation of random elements from $\mathcal{R}$, 2) evaluation 3) Insertion of significantly novel or well-performing agents into $\mathcal{R}$.

  In this work, we consider applications in which information about environment semantics is available, and can be appended to points in the agent's observation-action space or behavior space trajectories. An example of such a task is robot navigation in an environment where scene representations and annotations (\textit{e.g.} multimodal representations such as Concept-Fusion \cite{jatavallabhula2023conceptfusion}) have been previously estimated. This representation will be noted $\mathcal{I}_{s}$.

  In this context, our aim is to address two shortcomings of the repertoire $\mathcal{R}$:

  \begin{itemize}
    \item It is discrete and contains a finite number of agent/behavior pairs, and obtaining behaviors outside of this set is not straight-forward.
    \item The trajectories that are produced by this repertoire can not be customized beyond the \textemdash almost always low dimensional \textemdash behavior descriptor.
  \end{itemize}

  Our proposed solution, detailed in \S\ref{sec_proposed_method}, leverages Large Language Models (LLMs) along a causal transformer architecture to solve those problems jointly.
  
  Throughout the paper, Large Language Models will be referred to as $\Psi$, and their output will be written $\Psi(\xi(.))$ with $\xi(.)$ denoting a prompt. The notation $\xi(.)$ has been chosen to emphasize the fact that $\xi$ will often be a function of several variables, some of which are hand-designed and others automatically generated. When necessary for disambiguation, the role of the LLM in the pipeline will be made precise via a subscript, \textit{e.g.} $\Psi_{eval}$.

  \begin{figure*}[ht]
  \captionsetup[subfigure]{justification=centering}
      \centering
      \includegraphics[width=145mm,trim={2.0cm 2cm 1cm 1.0cm},clip]{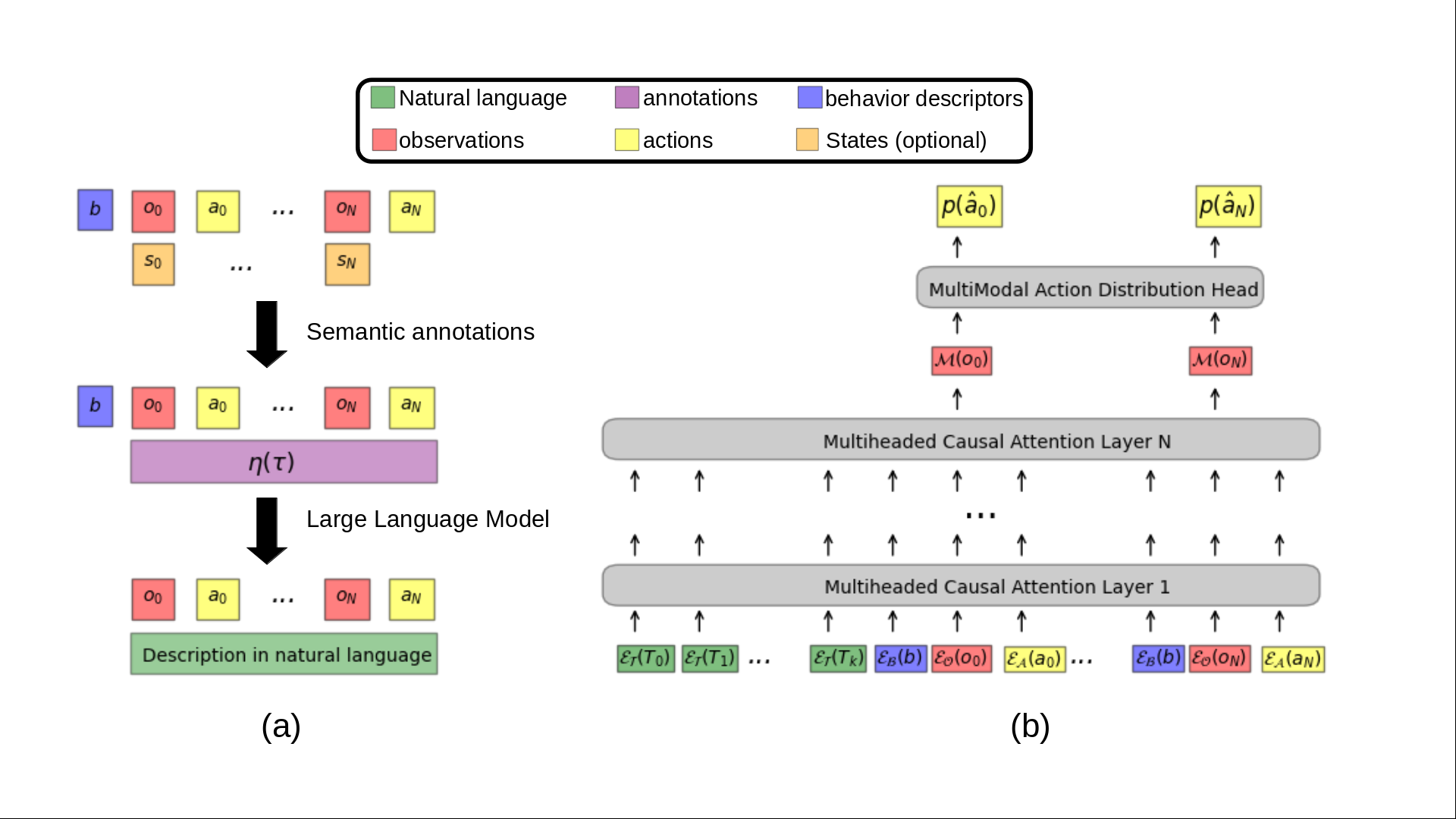}
      \caption{\small{High-level view of the proposed approach. Once a repertoire $\mathcal{R}$ has been populated by a Quality-Diversity method with agents along with their behavior descriptors, observation-action space \textemdash and optionally, state-space \textemdash \ trajectories, each of its elements is processed as in figure \textbf{(a)}: first, the semantic information $\mathcal{I}$ is used to annotate points in the trajectory. Then, those semantic annotations are passed to an LLM that is tasked with producing descriptions of those trajectories in natural language. These descriptions, which vary in terms of vocabulary, style and level of detail, are then added as meta data to the corresponding policies $\mathcal{R}$. \textbf{(b)} The model architecture. The transformer is conditioned on the text token embeddings $\mathcal{E}_{\mathcal{T}}$, which include positional embeddings. This is followed by the observation and action embeddings  $\mathcal{E}_{\mathcal{O}}, \mathcal{E}_{\mathcal{A}}$ of the entire trajectory that are intertwined with behavior descriptor conditioning embeddings $\mathcal{E}_{\mathcal{B}}$ at each timestep. Note that timestep embeddings are also included in $\mathcal{E}_{\mathcal{B}}, \mathcal{E}_{\mathcal{O}}, \mathcal{E}_{\mathcal{A}}$. At each timestep $i$, the $\mathcal{M}(o_i)$ embeddings that are output by the transformer for observation $o_i$ are mapped to a multimodal distribution $p(\hat{a}_i)$ over potential actions.}}
   \label{fig_high_level}
\end{figure*}

\section{Proposed method}
  \label{sec_proposed_method}

  The proposed approach distills a QD repertoire that has been enriched with natural language descriptions into a language and behavior conditioned policy modeled by a causal transformer. Figure \ref{fig_high_level} gives a high level view of our approach, the different components of which are detailed below.

\noindent \textbf{Enriching the repertoire with natural language.} QD archives traditionally contain pairs of policies and their associated behavior descriptors. In order to enable language grounding as well as supervision on the level of actions, each agent in the archive is first augmented with observation-action and optionally state-space trajectories. A function $\eta(\tau, \mathcal{I}_s)$ uses the semantic map $\mathcal{I}_s$, assumed available, to coarsely annotate each trajectory $\tau$ in the repertoire. The final natural language description for each $\tau$ is then obtained from an LLM $\Psi$ as $c=\Psi(\xi(\eta(\tau, \mathcal{I}_s),q))$, where $q$ is the hand-engineered component of the prompt $\xi$.

Those functions and the prompt $q$ are naturally problem-specific to variable extents. In our work, $\eta$ simply chooses points from $\tau$ at regular timesteps, and records the names or attributes of the areas and objects that the agent was close to at this time, as well as an indication of their relative position using relative cardinal and intercardinal directions. The string $q$ includes an explanation about $\eta$ and the environment layout, and instructs the LLM to map instances of $\eta(\tau, \mathcal{I}_s)$ to natural language requests, instructions or descriptions. See appendix \ref{sec_appendix_llm_interaction} for more details.

  \noindent \textbf{Prompt/BD conditioned Transformer.} The proposed model is based on a decoder-only causal transformer, similar to the Decision Transformer \cite{chen2021decision}. The term causal signifies that the transformer can't look into the future, \textit{i.e.} that the attention for each token is only computed based on those that precede it. More formally, noting $Q, K, V$ the query, key and value matrices \cite{vaswani2017attention}, attention-weighted embeddings are computed as 

\begin{equation}
  \mathrm{Softmax}(\frac{QK^t}{\sqrt{d}}+U)V
\end{equation}

where $U$ is an upper triangular matrix where non-zero elements tend to $-\infty$, and where $d$ is the size of the embeddings.

A sequence of tokens in this work is given by $\{T_1, ..., T_k, b, o_0, a_0, ..., b, o_l\}$, where the $T_i$ and $b$ are respectively tokens from the natural language and behavior conditionings, and $o_i, a_i$ denote observations and actions. Each of the different token types are then mapped to an embedding which also includes positional and timestamp embeddings:

\begin{equation}
  \begin{split}
    & \mathcal{E}_{\mathcal{T}}(T_i) \triangleq f_\mathcal{T}(T_i) + g_p\mathcal{T}(i)   \\
    & \mathcal{E}_{\mathcal{O}}(o_j) \triangleq f_\mathcal{O}(o_i) + g_t(j)   \\
    & \mathcal{E}_{\mathcal{A}}(a_j) \triangleq f_\mathcal{A}(a_i) + g_t(j))  \\
    & \mathcal{E}_{\mathcal{B}}(b) \triangleq f_\mathcal{B}(b) + g_t(j)   \\
  \end{split}
\end{equation}
 
 where $f_\mathcal{O}, f_\mathcal{A}, f_\mathcal{B}$ are MLPs, and $f_\mathcal{T}, g_p, g_t$ are learnable embedding lookup tables. Note that $g_p$ encodes positional information in the input sequence, while $g_t$ encodes timestamp information from the action-observation space trajectory. Details about the architecture and various hyperparameters can be found in appendix \ref{sec_appendix_implem}.

The embeddings $\mathcal{M}$ that are output by the transformer are then fed into different heads that together model a multimodal distribution over actions.

  \noindent \textbf{Handling multimodal action distributions.} Distributions over continuous actions in robotic navigation or manipulation tasks are often highly multimodal. For example, in a maze navigation task with the final position as the behavior descriptor, an agent can take a potentially infinite number of state-space trajectories. Therefore, metrics such as the MSE loss will fail to capture the underlying structure and can even lead to out of distribution action selection. Several recent works have proposed solutions based on diffusion models \cite{pearce2023imitating, ajay2022conditional, chi2023diffusion} and \cite{lynch2020learning}. In this work, we use the solution proposed by Shafiullah \textit{et al.} \cite{shafiullah2022behavior} for multimodal behavior cloning: the set of all action vectors present in the training split is partitioned into $K$ clusters using k-means, and each action $a$ is then decomposed as the sum of a cluster center and an offset, \textit{i.e.} $a=a_{cluster}+a_{offset}$. Assuming $d-$dimensional actions, we map each observation embedding $\mathcal{M}(o_i)$ to 1) a $1d$ array of length $k$ representing cluster indexes and 2) a $k\times d$ matrix $\Delta A$, such that its $i-th$ row is the $d-$dimensional offset that would need to be applied if $i$ were the true cluster index. The final loss function that we use to supervise our model is given by 

  \begin{equation}
    \frac{1}{BH}(-\sum_{i=1}^H (1-p_{true}^i)^{\kappa}\log(p_{true}^i) + \lambda ||a-\hat{a}||_2).
    \label{eq_supervision}
  \end{equation}

  The first term in the expression above is the focal loss function \cite{lin2017focal, shafiullah2022behavior}, where $p_{true}^i$ denotes the probability output by the softmax layer for the ground truth cluster at timestep $i$. The second term is the MSE between the target action $a$ and the estimated action $\hat{a}$ that results from taking the row from $\Delta A$ which corresponds to \textit{\textbf{the true cluster}}, hence the absence of normalization by $d$ in that term. The variables $B$ and $H$ in expression \ref{eq_supervision} denote batch size and horizon.

  \section{Evaluation metrics}
  \label{sec_how_to_eval}

  Let us consider a trained model $\pi$ with the given architecture that has been trained on a QD repertoire. We need to assess that this policy is capable of 1) reaching the target behavior descriptor and 2) following the textual prompt on which it has been conditioned.

 The first ability can be easily quantified via a distance metric on behavior descriptors. In this work, we simply use and MSE term $\mathcal{L}(b, b_{\pi|b,c} )$ where $b$ and $c$ respectively denote the behavior and the textual prompt.

Evaluating the second one is considerably more difficult as it requires integration between different modes. The process in charge of evaluating how closely the agent's behavior matches the prompt needs and understanding of both natural language and the geometric/semantic aspects of the scene. Let us write $S(\tau|c)\in [0,1]$ the (average) score attributed by human evaluators to a trajectory, knowing the prompt $c$. Arguably, such a measure can be considered as the gold standard for evaluating the behavior/prompt relationship in the majority of robotic navigation and manipulation tasks. However, $S$ is naturally expensive. We thus search for a surrogate function $S_{surrogate}(x) \in [0,1]$. 

  A possible choice is to learn a metric in the same spirit as the well-known Inception Score \cite{salimans2016improved} or the Fréchet Inception Distance \cite{heusel2017gans}. However, this, if done in a supervised manner, will require a significant amount of labeling, and if done in an unsupervised manner, \textit{e.g.} via contrastive learning, will still depend on the noisy labeling provided by the LLM during archive annotation, and will carry the biases and inaccuracy of that model. As a result, we leave that direction for future work. Instead, we consider the use of LLMs for that evaluation, casting the problem as a prompt engineering problem that maximizes a number of similarity/correlation metrics between human made evaluations and those made by a language model.

  More precisely, let us assume the availability of a small dataset $\mathcal{D}_{gt}\triangleq\{((\tau_i,c_i)\leftrightarrow S(\tau_i|c_i))\}_{i=1}^N$ of trajectories/contexts that have been issued scores by humans evaluators, and noting $\Psi_{eval}$ an LLM (or an ensemble of LLMs), let
  
  \begin{equation}
    S_{\Psi_{eval}}^q(\tau|c)\triangleq \Psi_{eval}(\xi(\tau,c,q))
  \end{equation}

  where $q$ is a request that instructs the LLM to evaluate how well $\tau$, \textit{i.e.} the agent's behavior matches the textual context $c$ provided to the transformer. Our aim is to find a prompt $q^*$ that steers the LLM towards producing scores that are close to those resulting from human evaluation. More formally, $q^*$ is an approximate solution to the multi-objective optimization problem

  \begin{equation}
    q^*=\argmax_q \{f_j(\{S_{\Psi_{eval}}^{q}(\tau_i|c_i),S(\tau_i|c_i)\}_{i=1}^N)\}_{j=1}^M
    \label{eq_maximize_correlation}
  \end{equation}

  where the $f_j$ are similarity/correlation criteria. Those that were used in this paper will be detailed in \S\ref{subsec_metric_correlation}.

\section{Experiments}
\label{sec_exp}

We first describe the benchmark environment that is used for evaluating the proposed method (\S\ref{subsec_semantic_maze}) before discussing a few qualitative results in section \ref{subsec_qualitative}. The metric that we use for quantitative evaluation is derived in section \ref{subsec_metric_correlation} and quantitative results are presented in section \ref{subsec_quantitative}.

Architecture, hyperparameter and dataset details are reported in appendix \ref{sec_appendix_implem}. Likewise, interactions with LLMs are detailed in appendix \ref{sec_appendix_llm_interaction}.

\subsection{Semantic Maze Environment}
\label{subsec_semantic_maze}

Simulated partially observable maze navigation is a popular benchmark for evaluating QD algorithms as well as the performance of RL approaches in sparse rewards settings \cite{pierrot2022diversity, duan2016benchmarking, doncieux2019novelty}. The environment that we have developed for the evaluation is an extension which adds semantic information to different areas of the maze, which itself is a POMDP with continuous states and actions. More precisely, the maze is divided in $3\times3$ tiles, with each tile being associated with a unique color. Furthermore, sprites representing household objects are placed in different areas. Figure \ref{fig_maze_env} (left) illustrates the benchmark environment.

\begin{figure*}[ht]
  \centering
  \captionsetup[subfigure]{justification=centering}
      \includegraphics[width=140mm,trim={0.0cm 0cm 0cm 0.0cm},clip]{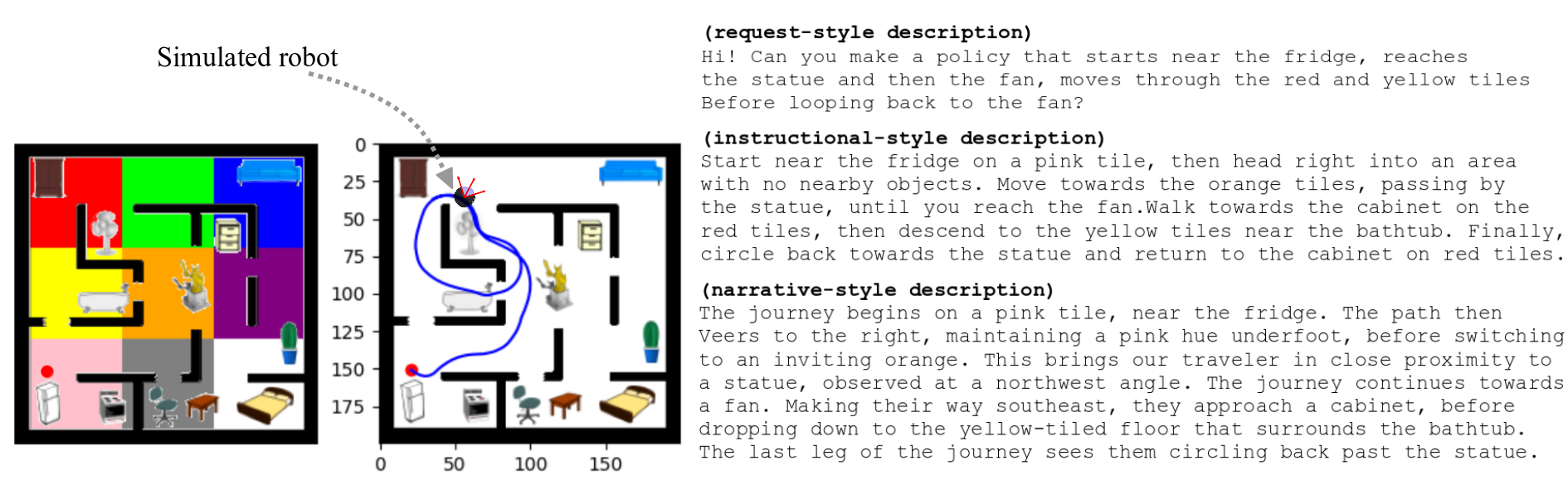}
      \caption{\small{\textbf{(left)} The proposed semantic maze navigation benchmark with continuous states and actions. The environment is partially observable, and sprites representing household objects as well as colored tiles have been added in order to provide static semantic information. The simulated robot, which is equipped with proximity sensors, starts at the red dot in the bottom left corner of the image. The policy controlling the robot is deployed in the environment during $400$ steps, and as is usual in navigation tasks in the QD literature, the $2d$ final position of the robot is recorded as its behavior descriptor. \textbf{(middle)} An example trajectory resulting from rolling out a policy from the repertoire. Colored tiles have been omitted for visibility. This trajectory is crudely annotated using the semantic information in the map. \textbf{(right)} Descriptions of the trajectory shown on the middle column generated by the LLM based on semantic annotations. For each trajectory, one of the three indicated styles in chosen randomly during training. }}
   \label{fig_maze_env}
\end{figure*}

The circular simulated robot starts at the red dot shown in figure \ref{fig_maze_env}(left) in the bottom left corner of the maze, and is equipped with five different sensors: three laser sensors that return the depth of an obstacle within a given range, and that are situated at angles $0, \frac{\pi}{2}, -\frac{\pi}{2}$ relative to the robots front. It is also equipped with two Boolean "bumpers", which return $1$ when the robot collides with a wall. We note that collisions are only possible with walls, and that the household objects do not constitute obstacles. The robot's action space is two-dimensional and similar to those of a turtlebot\footnote{https://www.turtlebot.com/}, which set the linear and angular velocity controls. As is usual in the QD literature, the final position of the robot was used as the behavior descriptor. We note that the environment's transition function is deterministic, and that the repertoire of policies that is built based on it does not require any particular uncertainty handling mechanisms. 

Figure \ref{fig_maze_env}(middle) illustrates a trajectory followed by a randomly selected agent from the repertoire, along with natural language descriptions generated by an LLM (figure \ref{fig_maze_env} (right)). As detailed in appendix \ref{sec_appendix_llm_interaction}, regularly selected points from the trajectory were first coarsely annotated with information about objects and tile colors that were closer to the agent at that timestep, as well as relative cardinal and intercardinal directions situating the agent relative to those objects. This annotation was then inserted in the prompt detailed in appendix \ref{sec_appendix_prompts_generation} in order to obtain high-level, natural language descriptions of the trajectory in three different styles: request, instruction and narrative. We alternated between the use of \texttt{gpt-3.5-turbo-0301} and \texttt{text-davinci-003} for this task, driven by the observed trade-off between precision/coherence and variation associated with those models. More details on the subject can be found in appendix \ref{sec_appendix_llm_interaction}.

We note that despite its deceptively simple appearance, the proposed benchmark is far from trivial: an agent going from the fridge to, say, the cactus, can pass through several object/colors with respect to which it can have different relative cardinal and intercardinal directions. This combinatorial nature, combined with the variation generated by the LLMs in their description as well as the multimodality of action distributions, makes an understanding of natural language and its correlations with various elements from the RL trajectory essential for generalization.

\begin{figure*}[ht]
    \centering
       \begin{subfigure}{0.44\textwidth}
        \includegraphics[width=\linewidth,trim={0.4cm 1.5cm 0.4cm 0.0cm},clip]{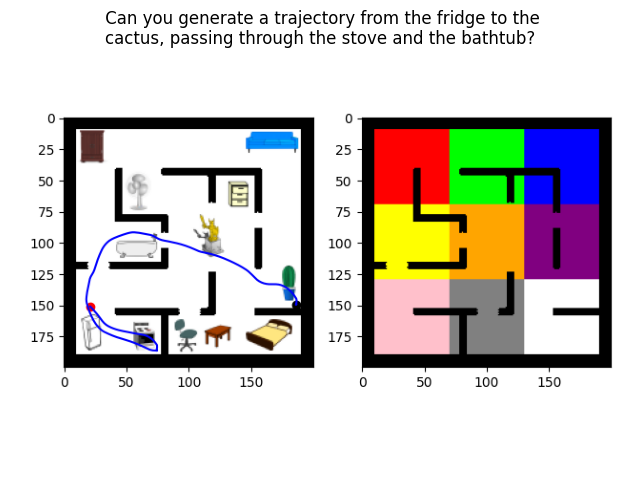}
        \caption{\scriptsize{Target behavior descriptor: (row=130, col=173)}}
    \end{subfigure}
    \begin{subfigure}{0.44\textwidth}
        \includegraphics[width=\linewidth,trim={0.4cm 1.5cm 0.4cm 0.0cm},clip]{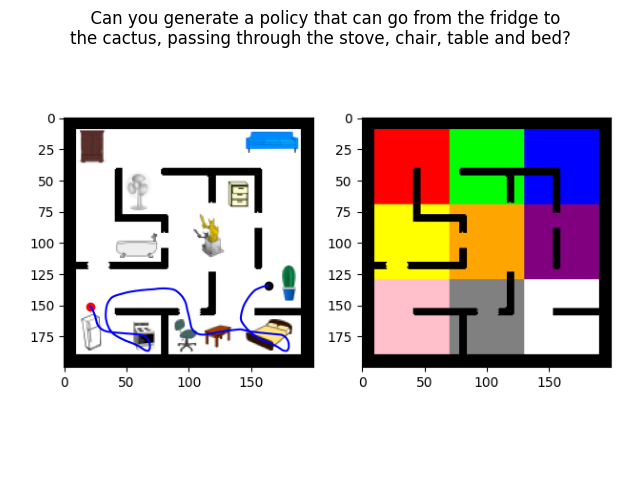}
        \caption{\scriptsize{Target behavior descriptor: (row=130, col=173)}}
    \end{subfigure}\\
    \begin{subfigure}{0.44\textwidth}
        \vspace*{0.5cm}
        \includegraphics[width=\linewidth,trim={0.4cm 1.5cm 0.4cm 0.0cm},clip]{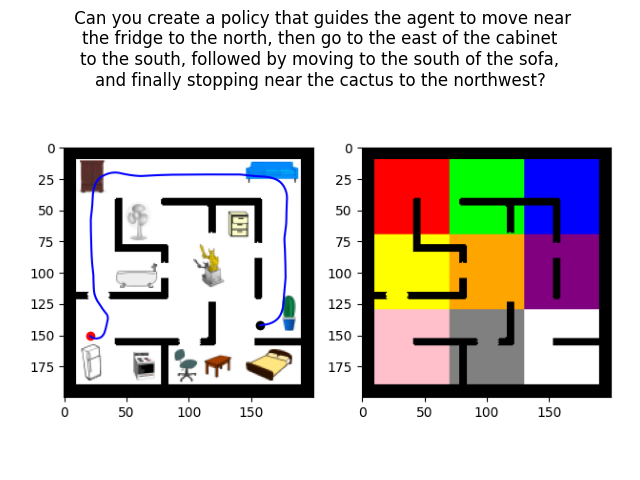}
        \caption{\scriptsize{Target behavior descriptor: (row=130, col=173)}}
    \end{subfigure}
      \begin{subfigure}{0.44\textwidth}
        \vspace*{0.35cm}
        \includegraphics[width=\linewidth,trim={0.4cm 0.8cm 0.4cm 0.0cm},clip]{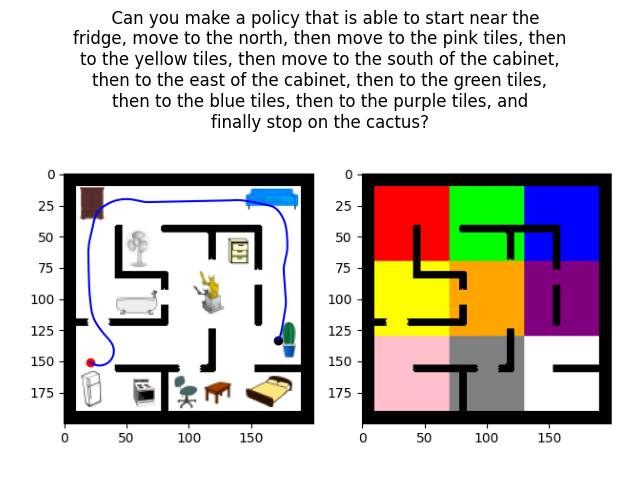}
        \caption{\scriptsize{Target behavior descriptor: (row=130, col=173)}}
    \end{subfigure}
    \begin{subfigure}{0.44\textwidth}
        \includegraphics[width=\linewidth,trim={0.4cm 0.0cm 0.4cm 0.0cm},clip]{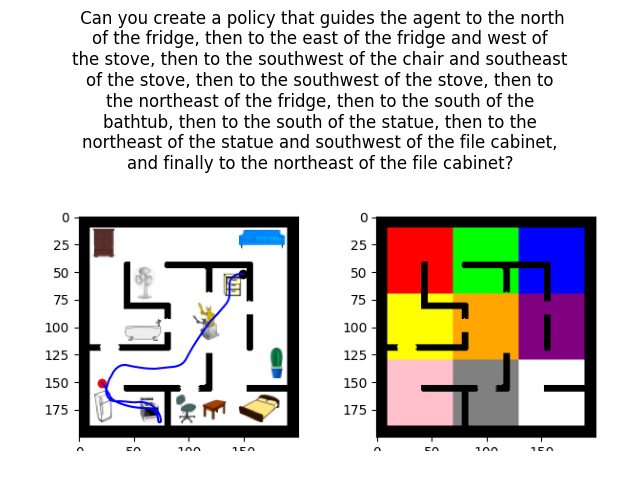}
        \caption{\scriptsize{Target behavior descriptor: (row=49, col=149)}}
    \end{subfigure}
    \begin{subfigure}{0.44\textwidth}
        \includegraphics[width=\linewidth,trim={0.4cm 0.0cm 0.4cm 0.0cm},clip]{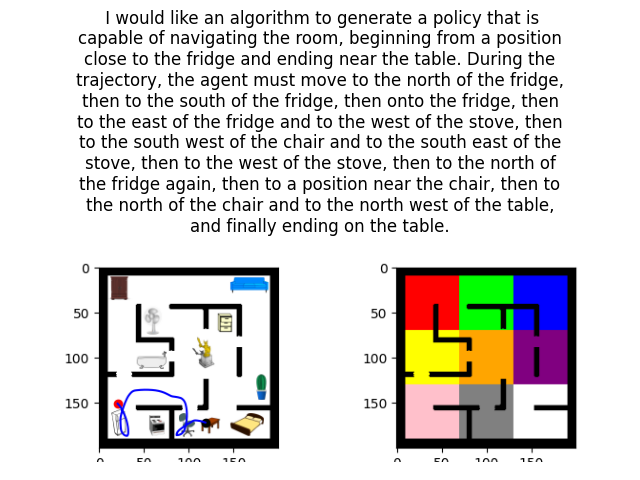}
        \caption{\scriptsize{Target behavior descriptor: (row=170, col=122)}}
    \end{subfigure}\\
    \caption{\small{Example trajectories generated by the conditioned transformer. The prompt and target behavior descriptors are respectively given above and below each figure, and for better visualization, colored tiles are displayed beside the trajectory instead of being overlaied. Note that (a,b,c,d) result from the same target behavior descriptor but with different textual prompts.}}
    \label{fig_qualitative}
\end{figure*}

\begin{figure*}[ht]
    \centering
       \begin{subfigure}{0.44\textwidth}
        \includegraphics[width=\linewidth,trim={0.4cm 1.5cm 0.4cm 0.0cm},clip]{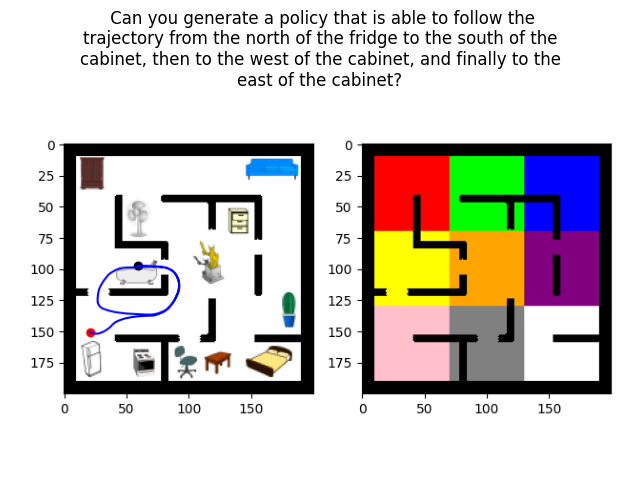}
         \caption{\scriptsize{Target behavior descriptor: (col=56, row=100)}}
    \end{subfigure}
    \begin{subfigure}{0.44\textwidth}
      \vspace*{0.35cm}
        \includegraphics[width=\linewidth,trim={0.4cm 2.2cm 0.4cm 0.0cm},clip]{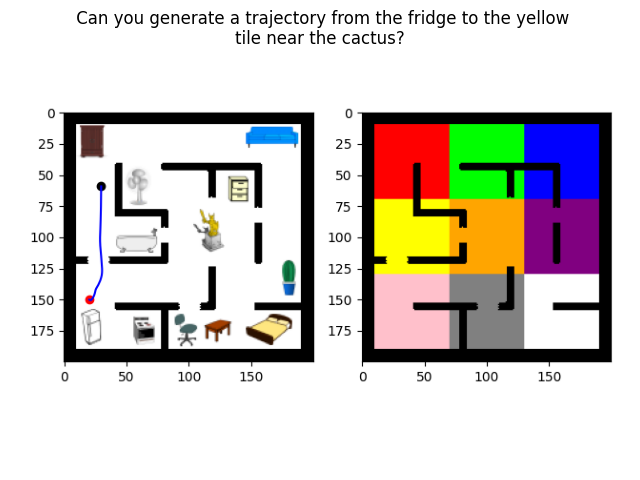}
        \caption{\scriptsize{Target behavior descriptor: (col=171, row=126)}}
    \end{subfigure}
    \caption{\small{Examples of generated solutions when behavior descriptors and textual prompt are incoherent. \textbf{(a)} In this example, the target behavior descriptor was centered on the bathtub, however, the prompt asks the policy to end near the cabinet (top left). \textbf{(b)} The target behavior descriptor lies within the purple area, but the prompt asks for something that is impossible as the yellow tile is not near the cactus.}}
    \label{fig_qualitative_screwups}
\end{figure*}

  \subsection{Qualitative results.}
  \label{subsec_qualitative}

  A few qualitative results are given in figure \ref{fig_qualitative}. In particular, figures \ref{fig_qualitative} (a, b, c, d), which result from conditioning on the same behavior descriptor (up to discretization error) but on different textual prompts, illustrate that the trained model has successfully learned to integrate all modalities. Figures \ref{fig_qualitative} (e,f) show results for conditioning on other target behavior descriptor and prompt combinations. Note that in all of those figures, the error in behavior, defined as the distance between the target behavior descriptor and the behavior descriptor reached in practice is less than $5\%$ of the diagonal of the maze.

  Of course, not all prompt and target behavior combinations result in satisfying trajectories. The network is in particular sensitive to 1) prompts that contradict the target behavior descriptor and 2) Prompts that are incoherent or ambiguous. Example trajectories in each of those situations are given in figure \ref{fig_qualitative_screwups}.

  In addition to the problems mentioned above, LLM generated labels are noisy and lead to sub-optimal supervision, and furthermore, test-time sampling from the multimodal action distribution introduces stochasticity. Hence the need for quantitative results, which we discuss in the following sections.

\begin{figure*}[ht]
    \centering
    \begin{subfigure}{0.24\textwidth}
        \includegraphics[width=\linewidth,trim={0.4cm 0cm 0.4cm 0.0cm},clip]{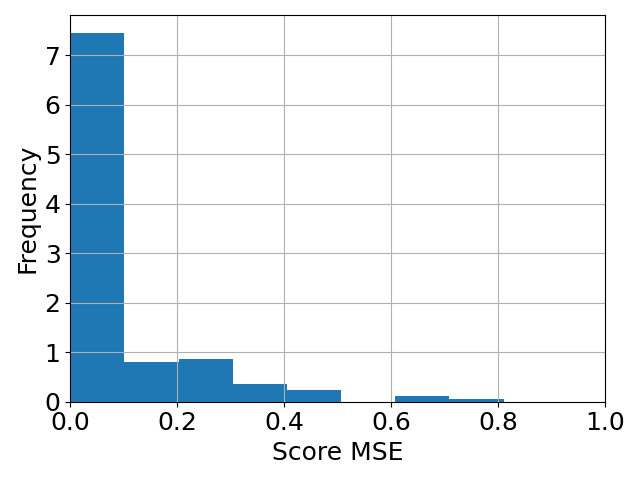}
        \caption{}
    \end{subfigure}
    \hfill
    \begin{subfigure}{0.24\textwidth}
      \includegraphics[width=\linewidth,trim={0.4cm 0cm 0.4cm 0.0cm},clip]{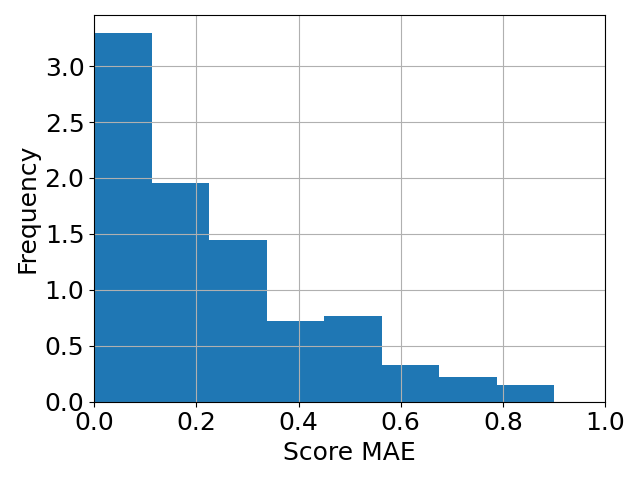}
      \caption{}
    \end{subfigure}
    \hfill
    \begin{subfigure}{0.24\textwidth}
      \includegraphics[width=\linewidth,trim={0.4cm 0cm 0.4cm 0.0cm},clip]{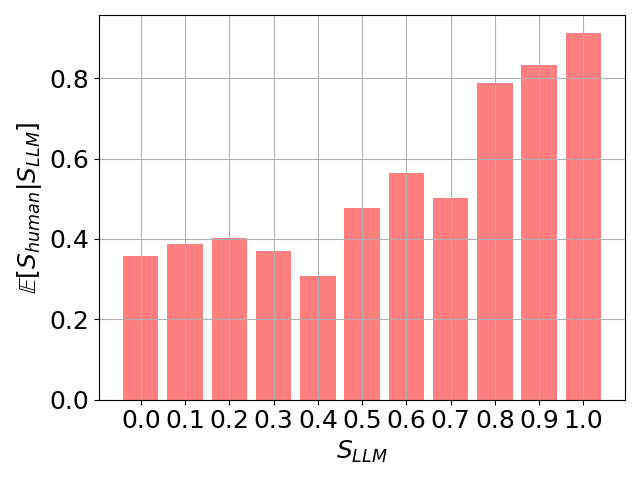}
      \caption{}
    \end{subfigure}
    \hfill
    \begin{subfigure}{0.24\textwidth}
      \includegraphics[width=\linewidth, trim={0.3cm 0cm 0.2cm 0.0cm},clip]{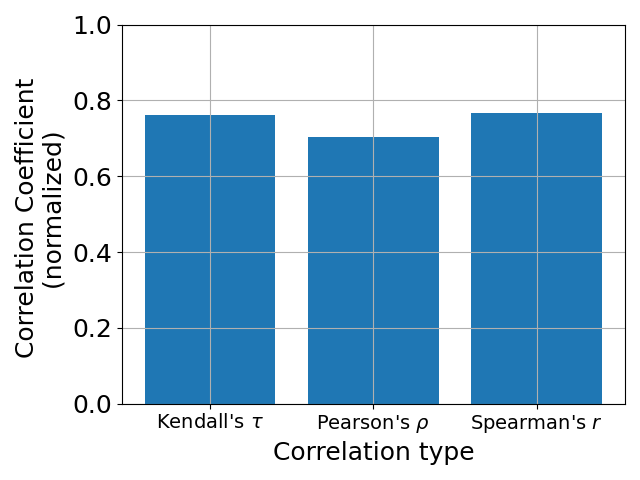}
      \caption{}
    \end{subfigure}
    \caption{\small{\textbf{(a)} Histogram of MSE between scores given by \texttt{gpt4-0314} and human subjects. \textbf{(b)} Histogram of MAE between scores given by \texttt{gpt4-0314} and human subjects. \textbf{(c)} $\mathbb{E}[S_{human}|S_{LLM}]$, \textit{i.e.} the expectation of the score given by humans knowing the score given by \texttt{gpt4-0314}. \textbf{(d)} Three complementary correlation measures (Kentall's $\tau$, Spearman's $r$ and Pearson's $\rho$) between scores given by \texttt{gpt4-0314} and human subjects. Note that those correlation values have been normalized in $[0,1]$.}}
    \label{correlation_fig}
\end{figure*}

  \subsection{Evaluation metric alignment with scores assigned by humans}
  \label{subsec_metric_correlation}

  The aim of this section is to adapt the LLM-based evaluation process described in section \ref{sec_how_to_eval}, which consists in finding a prompt that maximizes the $f_j$ of equation \ref{eq_maximize_correlation} based on a sample $\mathcal{D}_{gt}$ of trajectory-score pairs obtained from human annotators. We first describe the dataset and then discuss the choice of $f_j$, before presenting statistics on the surrogate $S_{\Psi_{eval}^{q^*}}$ obtained through optimizing the latter.

  \noindent\textbf{Alignment dataset.} For this section, human evaluators were asked to rate the correspondence between prompts and agent behaviors with a discrete score from $0$ to $10$, that were then normalized in $[0,1]$. As a result, for the rest of the paper, the functions $S$ and $S_{\Psi_{eval}}$ are assumed to take one of the $11$ regularly sampled values in $[0,1]$ for each trajectory:

  \begin{equation}
    \text{Im}(S)=\text{Im}(S_{\Psi_{eval}}^q)\triangleq \{0.0, 0.1, ..., 1.0\}.
  \end{equation}

  A number of $400$ trajectory-score pairs where gathered, where each score $S(\tau|c)$ was given by the average of the score assigned to the alignment between the trajectory and the prompt by five human evaluators. The trajectories were chosen randomly from results obtained with three different checkpoints, taken at different stages of training, in order to balance the number of poorly generated and well matching behaviors in the sample.

  \noindent\textbf{What $f_i$ to optimize?} A first natural choice is to consider the Mean Squared Error (MSE) between $S$ and $S_{\Psi_{eval}}^q$ and its more robust to outliers counterpart the Mean Absolute Error. However, those metrics do not provide inside into how well $S_{\Psi_{eval}}$ predicts $S$ on a per-value basis. For this reason, we also chose to monitor $||\mathbb{E}[S| S_{\Psi_{eval}}^q(\tau_j |c_j)] - S||$ over all possible $11$ values of $S$ during our trial and error based adjustment of the prompt $q$. As a complementary measure, the linear correlation between the two variables was also taken into account. However, as the two scoring processes must also be correlated in terms of their ranking \footnote{That is, if $S$ assigns a lower score to a trajectory-prompt pair $x$ than to another pair $y$, then we want $S_{\Psi_{eval}}$ to also rank $x$ lower than $y$.}, Kendal's $\tau$ as well as Spearman's correlation were additionally monitored. Those $f_j$ are summarized in table \ref{table_f_i}.

  \begin{table}[h]
    \centering
    \renewcommand{\arraystretch}{3.0}
    \begin{scriptsize}
    \begin{tabular}{|c|c|}
      \hline
      $f_0$    & -$\frac{1}{N}\sum_{i=1}^N||S(\tau_i|c_i)-S_{\Psi_{eval}}^q(\tau_i|c_i)||^2$ (MSE)   \\
      \hline
      $f_0$    & -$\frac{1}{N}\sum_{i=1}^N||S(\tau_i|c_i)-S_{\Psi_{eval}}^q(\tau_i|c_i)||$ (MAE)   \\
      \hline
      $f_2$    & -$\frac{1}{M}\sum_{j=1}^M||\mathbb{E}[S| S_{\Psi_{eval}}^q(\tau_j |c_j)] - S||$\\
      \hline
      $f_3$    & $\frac{(\#\text{concordant pairs}-\#\text{discordant pairs})}{\#\text{pairs}}$ (Kendall's $\tau$)  \\
      \hline
      $f_4$    & $\frac{\mathrm{Cov}(S,S_{\Psi_{eval}}^q)}{\sigma(S) \sigma(S_{\Psi_{eval}})}$ (Pearson correlation)  \\
      \hline
      $f_5$    & $\frac{\mathrm{Cov}(R(S),R(S_{\Psi_{eval}}^q))}{\sigma(R(S)) \sigma(R(S_{\Psi_{eval}}))}$ (Spearman correlation)  \\
      \hline
    \end{tabular}
    \end{scriptsize}
    \vspace*{0.3cm}
    \caption{\small{The functions used to evaluate the similarity/correlation between LLM generated trajectory-prompt alignment scores and those assigned by humans.}}
    \label{table_f_i}
  \end{table}

 \noindent\textbf{Statistics of the final prompt.} We adjusted the prompt through trial and error until we obtained a satisfactory $q^*$, given in appendix \ref{sec_appendix_prompts_eval}, which resulted in the statistics reported in figure \ref{correlation_fig}. The modes of the MSE and MAE histograms of figures \ref{correlation_fig}(a,b) are close to zero, and in both figures, most of the distribution's mass is concentrated on error values $<0.2$. Figure \ref{correlation_fig}(c) in particular shows that the scores produced by the LLM are a reasonable estimator of those produced by humans, as the $\mathbb{E}[S| S_{\Psi_{eval}}^{q^*}(\tau_j |c_j)]$ is consistently close to $S$. Finally, the results reported in figure \ref{correlation_fig}(d), which are normalized in $[0, 1]$, indicate moderate positive correlation between $S$ and $S_{\Psi_{eval}}^{q^*}$ both in terms of values and rankings.

  \subsection{Quantitative results}
  \label{subsec_quantitative}

\begin{figure*}[ht]
    \centering
       \begin{subfigure}{0.40\textwidth}
        \includegraphics[width=\linewidth,trim={0.0cm 0.0cm 0.0cm 0.0cm},clip]{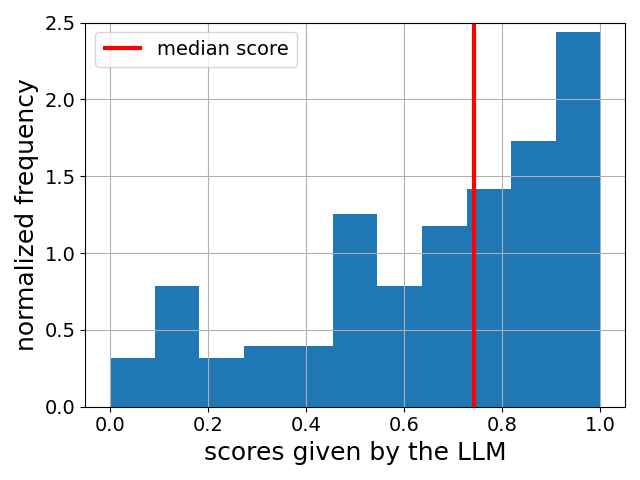}
         \caption{\scriptsize{histogram of scores given by the LLM (higher is better).}}
    \end{subfigure}
    \begin{subfigure}{0.40\textwidth}
        \includegraphics[width=\linewidth,trim={0.0cm 0.0cm 0.0cm 0.0cm},clip]{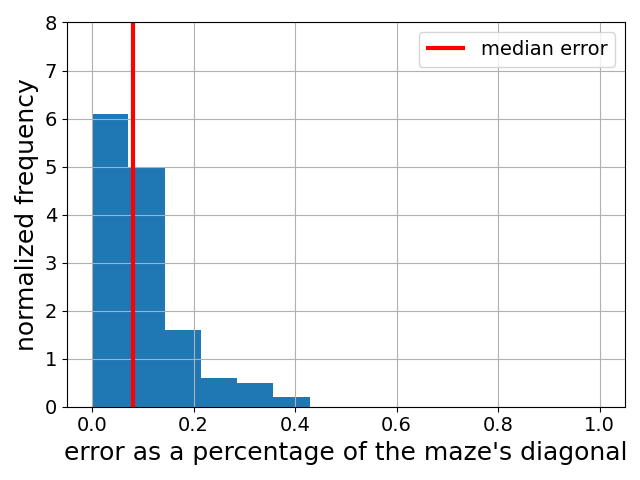}
        \caption{\scriptsize{histogram of behavior descriptor error (lower is better).}}
    \end{subfigure}
    \caption{\small{Histograms of \textbf{(a)} scores given to trajectories by the LLM and \textbf{(b)} behavior descriptor errors. Those results were obtained by selecting the best of five rollouts for each test example in terms of behavior descriptor error, before prompting the LLM for a score.}}
    \label{fig_quantitative}
\end{figure*}

  The test split on which the results in this section are based was composed of $900$ textual prompt and target behavior descriptor pairs. At test time, we simply sampled from the estimated multimodal action distribution. As this introduced stochasticity, five rollouts were made for each test example, and the trajectory with the lowest behavior descriptor error was selected for evaluation by the LLM using the metric derived in the previous section. Histograms of the resulting scores and behavior descriptor errors are given in figure \ref{fig_quantitative}. The median score ($\sim 0.74$) assigned by the LLM, as well as the median behavior descriptor error ($\sim 7\%$ of the maze's diagonal length) are also displayed.

We note the considerable stochasticity among the five rollouts: on average, each set of five rollouts \textemdash performed for the same example \textemdash have a mean and standard deviation of respectively $11.2\%$ and $14.6\%$ relative to the length of the maze's diagonal length. While this could partially be due to our action representation and sub-optimal test-time sampling, we conjecture that the most important factor leading to this error distribution is the significant amount of noise that is introduced by LLM generated descriptions both at training and test time. As mentioned in section \ref{subsec_qualitative}, the LLM often generates erroneous descriptions. Figure \ref{fig_qualitative_screwups} illustrated cases in which the descriptions where in contradiction with the target behavior descriptor or contained logical errors. However, those examples are obviously not exhaustive: the LLM generated descriptions sometimes contain hallucinated objects, miss several parts of the trajectory, or even reference information that is not exploitable by the agent, such as "The trajectory follows the information given by the user in the dictionary". Further analysis and mitigation of such sources of noise are however outside the scope of the current paper and are left for our future work.

\section{Related Works}
\label{sec_related_works}

\noindent\textbf{Generative Quality-Diversity.} Closely related to our work is the QD transformer \cite{mace2023quality} which is a decision transformer conditioned on target behavior descriptors. While we introduce a mechanism to model multimodal action distributions, their architecture is supervised using an MSE loss, which is not compatible with multimodal action distributions. Second, their policies are conditioned on target behavior descriptors whereas we extend this conditioning to both language and behavior descriptors. Finally, it should be noted that their experiments are performed in uncertain environments, where repeatability issues are mitigated using their proposed map-elites variant, map-elites low-spread. In contrast, our experiments are conducted in repeatable settings. However, it should be noted that uncertainty mitigating strategies such as map-elites low-spread, which concern the repertoire and not the model, are orthogonal to our work, and can be included as a preprocessing step.

The early work by Jegorova \textit{et al.}, uses a QD repertoire \cite{jegorova2020behavioral} to bootstrap the training of a GAN, which is only conditioned on target behavior descriptors. Furthermore, their focus is on the simpler case of open-loop policies represented by third order polynomials in time. In other words, policies that are sampled from the GAN \textit{are} the actions themselves, and can not be adjusted based on new observations.

\noindent\textbf{Offline RL.} Traditionally, the goal of offline RL has been to find a single return maximizing policy\cite{levine2020offline}. However, some recent works have shifted towards leveraging offline data to learn generative models of policies that are conditioned on return or skills \cite{ajay2022conditional}. Under a more general definition of offline RL that would include those conditional generative approaches, the model that we trained as part of the proposed method and which \textemdash much like QD transformer \cite{mace2023quality} \textemdash builds on decision transformers \cite{chen2021decision} can be viewed as a particular case of offline RL. In their paper where they propose the text-conditioned Frostbite benchmark, Putterman \textit{et al.} learn a transformer architecture similar to ours that is conditioned on textual prompts. While their architecture is similar to ours, they focus on discrete actions and do not condition on additional behavior descriptors. Furthermore, their textual prompts are less varied than the ones that are generated by our proposed LLM-based pipeline.

\noindent\textbf{Language grounding in robotics.} Early attempts at linking perception and actions to linguistic symbols can be traced back to the 1970s and were in general based on hand-crafted rule based systems \cite{winograd1972understanding}. During the following decades, probabilistic and learning approaches gradually became dominant \cite{mavridis2005grounded,tellex2011understanding, socher2014grounded, misra2016tell}. In particular, recent years have seen the emergence of a family of approaches related to goal-conditioned RL, where, as in our work, policies are conditioned on language \cite{stepputtis2020language, putterman2021pretraining}.

Since their introduction, foundation models \cite{bommasani2021opportunities, yang2023foundation}, which are large models that leverage web-scale data, have proven remarkably useful in problem areas related to language grounding. Examples of such models is CLIP\cite{radford2021learning} which learns joint text/image representations, and Large Language Models (LLMs) such as GPT-4 \cite{bubeck2023sparks}. Usage of foundation models ranges from the generation of goal states and reward shaping \cite{kapelyukh2023dall, kwon2023reward} to planning and control \cite{liang2023code,huang2022inner, ahn2022can, brohan2023rt}. Our work does not use LLMs for decision making, but only for data generation, in a manner that is similar to that of previous works \cite{liang2023code, schick2023toolformer}.

\section{Discussion}
\label{sec_discussion}

Our experiments demonstrated the capability or the proposed approach to generalize beyond the discrete repertoire, and to shape the behavior of the sampled agents according to prompts given in natural language. While we did not consider stochastic environments in our experiments, we note that uncertainty mitigation strategies\cite{mace2023quality} that are commonly used in the QD literature are complementary to our work, and can be included as a preprocessing step when building the repertoire.

We note the main limitations of our work: 1) Even though the policy is close-loop, the semantic information that it takes into account is based on pre-computed, frozen environment representations which will not capture the states of dynamic objects. 2) The clustering based modeling of action distributions introduces hyperparameters that can be difficult to fix in practice, and alternatives such as diffusion models might therefore be preferable. 3) The considerable variations in performance that were observed at test time are likely to be improved by considering more optimal sampling strategies \cite{holtzman2019curious}. 4) We did not quantify nor explicitly mitigate the amount of noise that was introduced by incoherent/illogical/hallucinated LLM generated descriptions. Addressing those limitations, as well as evaluating the proposed method in more realistic robotic environments constitute future directions that we hope to investigate.

\acknowledgments{This work was supported by the European Union’s Horizon EU.2.4 Digital, Industry and Space Research and Innovation Program through FET Project PILLAR\footnote{\url{https://pillar-robots.eu/}}(Grant Agreement No. 101070381).}


\bibliography{example}  

\appendix
\section{Implementation details}
\label{sec_appendix_implem}

\noindent\textbf{Architecture details and hyperparameters.} Our transformer architecture is based on GPT-2. Specifically, our code \footnote{\url{https://github.com/salehiac/LanguageGroundedQD}} extends nanoGPT \footnote{\url{https://github.com/karpathy/nanoGPT}}, which is a language model, to take as input RL trajectories and behaviors, and to output distributions over actions instead of probabilities over next tokens. This naturally requires the addition of observations, actions and behavior descriptor embedding layers. Furthermore, the multimodal action prediction scheme detailed in \S\ref{sec_proposed_method} requires k-means based action space partitioning as well as additional prediction heads. Table \ref{table_arch} summarizes those details and the relevant hyperparameters.

As for text tokenization, it should be observed that the environment used for experimental validation requires a much smaller vocabulary size for trajectory description than what is necessary in general NLP tasks. Therefore, we trained a specialized tokenizer on the corpus defined by the LLM generated trajectory descriptions, which resulted in a vocabulary size of about $1500$.

\noindent\textbf{Training.} The transformer was trained on a repertoire of $\sim 16k$ trajectories, that were each associated to an LLM generated description. The validation and test splits were each composed of $\sim 900$ trajectories. Those trajectories were obtained through combining five archives generated by vanilla Novelty Search\cite{lehman2011abandoning}, and the descriptions were obtained using the LLMs described in appendix \ref{sec_appendix_llm_interaction}.

MLPs used in the attention layers do not have any biases, while those that are used for embeddings and in prediction heads include bias parameters. All parameters except bias parameters are subject to L2 regularization with a weight of $1e-1$. We used dropout with $p=0.1$ only on the output of attention layers. Training was done using the AdamW optimizer \cite{loshchilov2017decoupled}, with parameters $\beta_1=0.9, \beta_2=0.95$. An initial warmup of $2000$ step led to the maximum learning rate of $3e-5$ that was then decayed using a cosine LR scheduler. The batch size was set to $30$ in our experiments.

We selected the final model based on the value of a modified version of the loss function (equation \ref{eq_supervision}) on validation data: instead of computing the second term based on the offset predicted for the ground truth cluster, we used the offset that was predicted for the cluster that was output by the cluster prediction head.

\noindent\textbf{Test time sampling.}  For testing, we simply sampled actions from the multimodal distribution defined by the k-means clusters.

\begin{table}[h]
    \centering
    \begin{tiny}
    \begin{tabular}{|c|c|c|}
        \hline
        \multirow{6}{*}{Transformer architecture}
        & embedding size (all modalities) & 360 \\
        \cline{2-3}
        & \#attention layers & 2 \\
        \cline{2-3}
        & \#attention heads per layer & 4 \\
        \cline{2-3}
        & context size & 1435 (226 for text tokens and the rest for the trajectory) \\
        \cline{2-3}
        & attention block: \#MLP hidden layer & 1 \\
        \cline{2-3}
        & attention block: MLP hidden layer dims & $4 \times$ \#embedding\_size \\
        \hline
        \multirow{2}{*}{Embedding layers} 
        & \#MLP hidden layers & 2 \\
        \cline{2-3}
        & \#MLP hidden layers dims & $2 \times$ \#embedding\_size \\
        \hline
        \multirow{1}{*}{k-means clustering}
        & \# clusters per action dimension & 4 for the first action dimension, 4 for the second.\\
        \hline
        \multirow{4}{*}{Prediction heads} 
        & \#prediction heads & 3 (two for per dimension action cluster index prediction, one for offset prediction) \\
        \cline{2-3}
        & \#MLP hidden layers & 2 \\
        \cline{2-3}
        & \#MLP hidden layers dims: cluster prediction & 64 \\
        \cline{2-3}
        & \#MLP hidden layers dims: offset prediction & 1024 \\
        \hline
        \multirow{1}{*}{\# total parameters} 
        & \ \  & 8.38M \\
        \hline
    \end{tabular}
    \end{tiny}
    \caption{\small{Architecture related hyperparameters used in our experiments.}}
    \label{table_arch}
\end{table}

 \section{Interactions with LLMs}
\label{sec_appendix_llm_interaction}

We augmented each trajectory in the repertoire $\mathcal{R}$ corresponding to the maze navigation environment with semantic and positional information. More precisely, points from the trajectory were selected at regular time intervals and annotated with information about the objects and color tiles that were closer to them, as well as with relational information. An example of such an annotation is given in figure \ref{fig_traj_annotation}. These annotations were used for the generation of natural language descriptions of trajectories, as well as for the evaluation of the alignment between the textual conditioning and the policies behavior.

\subsection{Mapping semantic annotations to natural language descriptions/requests/instructions}
 \label{sec_appendix_prompts_generation}

 \begin{figure}[h]
     \centering
     \begin{subfigure}{0.9\textwidth}
       \begin{Verbatim}[frame=single, fontsize=\tiny]
[
  {'timestep': 0, 'pos': [20.6, 49.5], 'semantics': ['to the north  of fridge'], 'colors': 'pink'},
  {'timestep': 40, 'pos': [23.0, 72.8], 'semantics': [], 'colors': 'yellow'}, 
  {'timestep': 80, 'pos': [16.6, 107.1], 'semantics': [], 'colors': 'yellow'},
  {'timestep': 120, 'pos': [15.6, 150.1], 'semantics': ['to the south west of cabinet'], 'colors': 'red'}, 
  {'timestep': 160, 'pos': [29.5, 179.0], 'semantics': ['to the  east of cabinet'], 'colors': 'red'}, 
  {'timestep': 200, 'pos': [70.1, 184.2], 'semantics': [], 'colors': 'green'}, 
  {'timestep': 240, 'pos': [113.8, 184.3], 'semantics': [], 'colors': 'green'}, 
  {'timestep': 280, 'pos': [157.6, 184.0], 'semantics': ['to the  west of sofa'], 'colors': 'blue'}, 
  {'timestep': 320, 'pos': [180.2, 164.5], 'semantics': ['to the south east of sofa'], 'colors': 'blue'}, 
  {'timestep': 360, 'pos': [182.8, 122.3], 'semantics': [], 'colors': 'purple'}, 
  {'timestep': 399, 'pos': [183.1, 79.5], 'semantics': ['to the north  of cactus'], 'colors': 'purple'}
]
 \end{Verbatim}
     \end{subfigure}
   \caption{\small{Example of semantic trajectory annotation that were included in the prompts given to LLMs.}}
   \label{fig_traj_annotation}
 \end{figure}

 We used two models released by OpenAI for this task: \texttt{gpt-3.5-turbo-0301} and \texttt{text-davinci-003}. The reason for this was that we observed that the former was much more capable in terms of introducing variations in style and vocabulary, but was overall less consistent in terms of logical reasoning and the generation of coherent instructions than \texttt{text-davinci-003}. We generated about $60\%$ of the archive annotations with \texttt{text-davinci-003} and the other $40\%$ was generated using \texttt{gpt-3.5-turbo-0301}. Prompts that were used with the latter are given in figure \ref{fig_prompt_gen}. We used similar prompts with \texttt{text-davinci-003}. In both cases, we did not observe significant improvements with few-shot or chain of thought prompting, and therefore kept the simplest (and less expensive) form reported in the aforementioned figure.

 Unlike the \texttt{gpt-4-0314} model that was used for model evaluation (see section \ref{sec_how_to_eval} and appendix \ref{sec_appendix_prompts_eval}), we found that additional information such as an ascii representation of the layout and the specification of object coordinates dramatically lowered the ability of \texttt{gpt-3.5-turbo-031} and \texttt{text-davinci-003} to generate coherent outputs.

\begin{figure}[h]
  \begin{tiny}
  \centering
  \begin{subfigure}{0.85\textwidth}
  \begin{lstlisting}[firstnumber=1]
  System: You are a useful assistant.
  User: Let us consider a square of size 200x200. The origin is fixed at the bottom left corner, and the x,y axes are respectively horizontal and vertical (with x looking towards the right, i.e. east, and y looking upwards, i.e. north). Let us define a python dictionary to represent point that have been sampled from a 2d trajectory in that square. The dictionary will have the following keys: dict_keys(['timestep', 'pos', 'semantics', 'colors']). Here is the explanation for each: 1) The complete trajectories are composed of N points, and each has a timestep t_i (with i ranging from 0 to N-1).  The 'timestep' key corresponds to the t_i of the sampled point. 2) the 'pos' key is for the 2d position of the point, expressed as (x,y) in the coordinate frame defined above. 3) The square actually represents a room, where there are several objects. The 'semantics' gives information about objects to which the points are close (name of objects, and agent position w.r.t those, e.g. to the east of the cactus. Note that east, west, north, south directions are relative to objects, not relative to the agent. Also, east=right, west=left, north=up, south=down). 4) The room which is represented by the 200x200 square also has tiles of different colors in different areas. The 'colors' key gives information about the tile color where the 2d point is. 
 \end{lstlisting}
    \caption{\small{The conditioning common to all \texttt{gpt-3.5-turbo-0301} prompts that were used to map semantic annotations to natural language descriptions/instructions/requests. Prompts used with the \texttt{text-davinci-003} where similar, with minor adjustments such as removing the system conditioning.}}
  \end{subfigure}

    \begin{subfigure}{0.85\textwidth}
      \begin{lstlisting}[firstnumber=3]
    User: Your task is to ask an algorithm to generate an agent capable of going through the given trajectory. An example of such a request could be 'Can you make a policy that is able to go near the A, then to the east of the B, before going to C and then stopping near D?', where A, B, C, D are extracted from the python dictionary described above (you should never use those letters without replacing them with relevant info from the dict). Another example could be 'Hi! My name is <insert_human_name_here>! I want you to make a network that can go from the fridge to A?' or 'Can you generate a trajectory from A to B that goes through C?'. Any combination of objects and tile colors is acceptable in your request as long as it matches the trajectory described in the python dict. You can also arbitrarily replace the last position/object/tile at the end of the trajectory with '<DESCRIPTOR>' if you want. NOTE: you must NEVER mention the python dictionary in the description. For example, 'Can you generate a policy that is able to follow the trajectory described in the given python dictionary?' is unacceptable. Please do NOT use any numerical values such as coordinates, and do NOT use timestep values, at all. The trajectory will be given after the tag [TRAJECTORY], and I want you to write your request after [REQ].
 \end{lstlisting}
      \caption{\small{Prompt used for request style text generation.}}
  \end{subfigure}

  \begin{subfigure}{0.85\textwidth}
      \begin{lstlisting}[firstnumber=3]
      User: Your task is to describe such trajectories with text, without any numerical values. So , NO coordinates, No timestep values. Please try to be concise. The trajectory will be given after the tag [TRAJECTORY], and I want you to write the description after [DESCR].
 \end{lstlisting}
    \caption{\small{Prompt used for request style text generation.}}
  \end{subfigure}

    \begin{subfigure}{0.85\textwidth}
      \begin{lstlisting}[firstnumber=3]
      User: Your task is to describe such trajectories with text, without any numerical values. Note that if there is no significant motion during the entire trajectory, it is acceptable to give a very concise description such as 'stay near the starting point'. Also, try to make those descriptions instructional, as if you were trying to guide an agent. Important: please be concise. The trajectory will be given after the tag [TRAJECTORY], and I want you to write the description after [DESCR].
 \end{lstlisting}
      \caption{\small{Prompt used for request style text generation.}}
  \end{subfigure}
    
    \caption{\small{Prompts used to generate natural language descriptions of trajectories with \texttt{gpt3-turbo-0301} and \texttt{text-davinci-003}.}}
    \label{fig_prompt_gen}
  \end{tiny}
\end{figure}

\subsection{LLM based Evaluation of behavior and textual prompt alignment}
 \label{sec_appendix_prompts_eval}

 We used the \texttt{gpt-4-0314} model via OpenAI's api for this task, using the prompt given in figure \ref{fig_gpt4_eval}. We found that asking the model to answer questions relevant to its task before reviewing those and detailing its step by step reasoning improved its ability of generating scores that were close to those of human evaluators. Furthermore, we found that specifying the environment layout in ascii improved the produced scores. We note that we could not obtain results on par with those of that model using \texttt{gpt-3.5-0301} or \texttt{text-davinci-003}.

\begin{figure}[h]
  \centering
  \begin{subfigure}{0.9\textwidth}

  \begin{Verbatim}[frame=single, fontsize=\tiny]

Let us consider a square of size 200x200. The origin is fixed at the bottom left corner, and the x,y axes
are respectively horizontal and vertical (with x looking towards the right, i.e. east, and y looking upwards,
i.e. north). Let us define a python dictionary to represent point that have been sampled from a 2d trajectory 
in that square. The dictionary will have the following keys: dict_keys(['timestep', 'pos', 'semantics', 
'colors']). Here is the explanation for each: 1) The complete trajectories are composed of N points, and each
has a timestep t_i (with i ranging from 0 to N-1).  The 'timestep' key corresponds to the t_i of the sampled 
point. 2) the 'pos' key is for the 2d position of the point, expressed as (x,y) in the coordinate frame 
defined above. 3) The square actually represents a room, where there are several objects such as a 
fridge, a chair, a cactus and so on. The 'semantics' gives information on objects to which the point 
are close (name of objects, and where the agent is situated w.r.t those objects, e.g. to the east or 
north of the cactus, etc). 4) The room which is represented by the 200x200 square also has tiles of
different colors in different areas. The 'colors' key gives information about the tile color where the 
2d point is.

******* 
Colored tiles: the 200x200 room is divided in 9 tiles, each of size approximately 66x66. From top to bottom, 
the layout is given below.

      Red     | Green  | Blue
      -------------------------
      Yellow  | Orange | Purple
      -------------------------
      Pink    | Gray   | White

*******
Here is a list of the objects present in the environment, their coordinates and the color of the tiles they
are on:

    ============= table 1============
    - cactus, (180, 27), purple
    - fan, (60, 140),  red/yellow
    - sofa, (167,180), blue
    - bed, (164,26), white
    - fridge, (22,28), pink
    - wood cabinet, (23,176), red
    - chair, (98,25), gray
    - statue, (120,105), orange
    - file cabinet, (140,138), orange/green
    - bathtub, (58,97), yellow
    - table, (123, 25), gray
    - stove, (64,25), pink

*******
You will receive a natural language description after the [DESCR] tag. This description has been given as a 
command to an agent, which has tried to execute it in the environment. The agent's resulting trajectory will
be given to you after the tag [DICT], using the dictionary format described above. Your task is to evaluate
how well the agent has been able to follow the instructions. More precisely, your task will be composed of
those steps:

A) Answer those questions:

    Q1. What was the description asking the agent to reach?
    Q2. What did the agent reach?
    Q3. What was the final destination that the agent was expected to reach?
    Q4. Did the agent reach that final destination? At which point in the dictionary?
    Q5. How far is the agent's final state from the expected final state?
    Q6. What objects/colors was the agent asked to visit, reach, pass by or encounter?
    Q7. What objects/colors did the agent actually visit, reach, pass by or encounter? How many were extra?
        How many did it miss?

    Note that in the table 1 above, a correspondance between colors, coords and objects has been given.

B) Assume that your end goal is to write a numerical score in [0,1] that rates how well the agent's behavior
matches the desired description. How would you proceed here? Please write your reasoning here.

C) Using what you know so far, write a numerical score in [0,1] that rates the agent's performance. You must 
   always come to a numerical decision, no matter how difficult the task.

*******

FORMATTING RULE: you must write the final score as 'score==<floating_value>'

*******
 \end{Verbatim}
  \end{subfigure}
    \caption{\small{The initial prompt given to \texttt{gpt-4-0314} for the evaluation of trajectory/textual conditioning alignment.}}
    \label{fig_gpt4_eval}
\end{figure}

\end{small}
\end{document}